\documentclass[12pt, a4paper]{article}

\usepackage[linesnumbered,ruled,vlined]{algorithm2e}
\usepackage{caption}
\usepackage{subfig}
\usepackage{amssymb}
\usepackage{authblk}
\usepackage[utf8]{inputenc}
\usepackage{hyperref}
\usepackage{subfig}
\usepackage{tikz}

\title{Segmentation of scanning electron microscopy images from natural 
rubber samples with gold nanoparticles using starlet wavelets\footnote{\textit{Published
in Microscopy Research and Technique (January 2014).} The final publication is available at
\url{http://dx.doi.org/10.1002/jemt.22314}.}}

\author[a]{Alexandre Fioravante de Siqueira\footnote{Corresponding author. Phone: +55(19)3521--5362.\\
                                           \href{alexandredesiqueira@programandociencia.com}{alexandredesiqueira@programandociencia.com}}}
\author[b]{Flávio Camargo Cabrera\footnote{\href{flavioccabrera@yahoo.com.br}{flavioccabrera@yahoo.com.br}}}
\author[c]{Aylton Pagamisse\footnote{\href{aylton@fct.unesp.br}{aylton@fct.unesp.br}}}
\author[b]{Aldo Eloizo Job\footnote{\href{job@fct.unesp.br}{job@fct.unesp.br}}}

\affil[a]{DRCC -- Departamento de Raios Cósmicos e Cronologia,
          IFGW -- Instituto de Física ``Gleb Wataghin'',
          UNICAMP -- University of Campinas,
          Rua Sérgio Buarque de Holanda, 777, 13083-970,
          Campinas, São Paulo, Brazil}
\affil[b]{DFQB -- Departamento de Física, Química e Biologia,
          FCT -- Faculdade de Ciências e Tecnologia,
          UNESP -- Univ Estadual Paulista,
          Rua Roberto Simonsen, 305, 19060-900,
          Presidente Prudente, São Paulo, Brazil}
\affil[c]{DMC -- Departamento de Matemática e Computação,
          FCT -- Faculdade de Ciências e Tecnologia,
          UNESP -- Univ Estadual Paulista,
          Rua Roberto Simonsen, 305, 19060-900,
          Presidente Prudente, São Paulo, Brazil}

\begin{document}

\maketitle

\clearpage

\begin{abstract}
Electronic microscopy has been used for morphology evaluation of
different materials structures. However, microscopy results may be
affected by several factors. Image processing methods can be used to
correct and improve the quality of these results. In this paper we
propose an algorithm based on starlets to perform the segmentation of
scanning electron microscopy images. An application is presented in
order to locate gold nanoparticles in natural rubber membranes. In this
application, our method showed accuracy greater than 85\% for all test
images. Results given by this method will be used in future studies, to
computationally estimate the density distribution of gold nanoparticles
in natural rubber samples and to predict reduction kinetics of gold
nanoparticles at different time periods.

\vspace{0.2cm}
\noindent\textbf{Keywords:} Image Processing, Gold Nanoparticles,
Natural Rubber, Scanning Electron Microscopy, Segmentation, Wavelets
\end{abstract}

\section{Introduction} 
\label{INTRODUCTION}

Recently, electronic microscopy has been widely used for morphology 
evaluation of different materials' micro and nanostructures, e.g. natural 
rubber/gold nanoparticles membranes\cite{CABRERA2013}, spray layer-by-layer 
films\cite{AOKI2012}, anisotropic metal structures\cite{DUCHENE2013}, 
Langmuir-Blodgett monolayers\cite{GUERRERO2010}, among others. However, 
microscopy results may be affected by factors such as equipment 
configuration, image resolution, external conditions, such as noise level 
and power grid stability, and even the analyzed material, which can 
present surface degradation depending on the energy level of the applied 
electron beam.

Images obtained by electronic microscopy do not often present good 
quality, especially when evaluated at nanoscale structural levels, where 
the parameters discussed above become even more influential. Image processing 
methods can be used when the image is not suitable for analysis, or when 
the characterization needs to be automated. Digital image processing is a 
powerful and well-established set of techniques, such as filtering, restoration, 
reconstruction, object recognition and segmentation. This last processing 
method separates an image into its constituent objects\cite{GONZALEZ2008}, recognizing 
a certain region of interest.

\subsection{Related work}
\label{RELWORK}

Image segmentation is often used in microscopy images that are currently 
segmented to obtain features such as number of cells\cite{SUI2013}, separation of 
overlapped particles\cite{MALLAHI2013} and skull-stripping of mouse brain \cite{LIN2013}.

There are a number of commercial and free software available for these 
processing \cite{USAJ2011}; however, these tools are focused on some images, 
and cannot be extended to analysis of other ones \cite{USAJ2011,SHITONG2006}. 
Furthermore, segmentation of nontrivial images is one of the most challenging 
tasks in image processing: its accuracy determines the success of 
computational analysis procedures \cite{GONZALEZ2008}.

\subsection{Proposed approach}

In this paper we propose an algorithm based on starlets to perform the 
segmentation of scanning electron microscopy images. This algorithm is 
applied in order to segment gold nanoparticles incorporated on natural 
rubber membranes (NR/Au). Starlets are isotropic undecimated wavelet 
transforms, well adapted to astronomical data, where objects present 
isotropy in most cases \cite{STARCK2006}. The proposed approach consists 
of applying the starlet transform in a sample image to obtain its detail 
decomposition levels. These levels are used to identify and separate 
background elements and noise from interest regions displayed on the input 
image.

Natural rubber samples were obtained from Hevea brasiliensis latex (RRIM 600 
clones) by casting method, and gold nanoparticles were added by in situ 
chemical reduction. These samples were used for chemistry analysis and 
ultrasensitive detection by Raman spectroscopy in the construction of flexible 
SERS and SERRS substrates\cite{CABRERA2012}, and also in the study of the NR/Au influence 
on the physiology of \textit{Leishmania braziliensis} protozoans \cite{BARBOZAFILHO2012}. 
Results presented in this study will be used as basis for the prediction of 
kinetics of gold nanoparticles reduced at different time periods in samples 
of natural rubber, and also in the density distribution study of gold 
nanoparticles in natural rubber.

The remainder of the paper is organized as follows. Section \ref{FORMULATION} introduces a brief 
background on starlet wavelets, the main tool of this method, and an overview 
of the proposed algorithm. Section \ref{EVALUATION} presents the image dataset used to test 
the algorithms, as well as the evaluation method to test the algorithm 
performance. Next, Section \ref{EXPERIMENTALRESULTS} exhibits the experimental results from algorithm 
application on elements of the image dataset. Also, we discuss the performance 
of the method. Finally, in Section \ref{CONCLUSION} we report our final 
considerations about this study.

\section{Formulation} 
\label{FORMULATION}

\subsection{Background on starlet transform}

Starlet is an undecimated (or redundant) wavelet transform based on the 
algorithm ``à trous'' (with holes) from Holschneider\cite{HOLSCHNEIDER1990} and 
Shensa \cite{SHENSA1992}. This wavelet is well suited to analyze
astronomical\cite{STARCK2006,STARCK2010} 
or biological\cite{GENOVESIO2003} images, that usually contain isotropic objects.

\begin{figure*}[hbt]
\centering
\includegraphics[width=10cm]{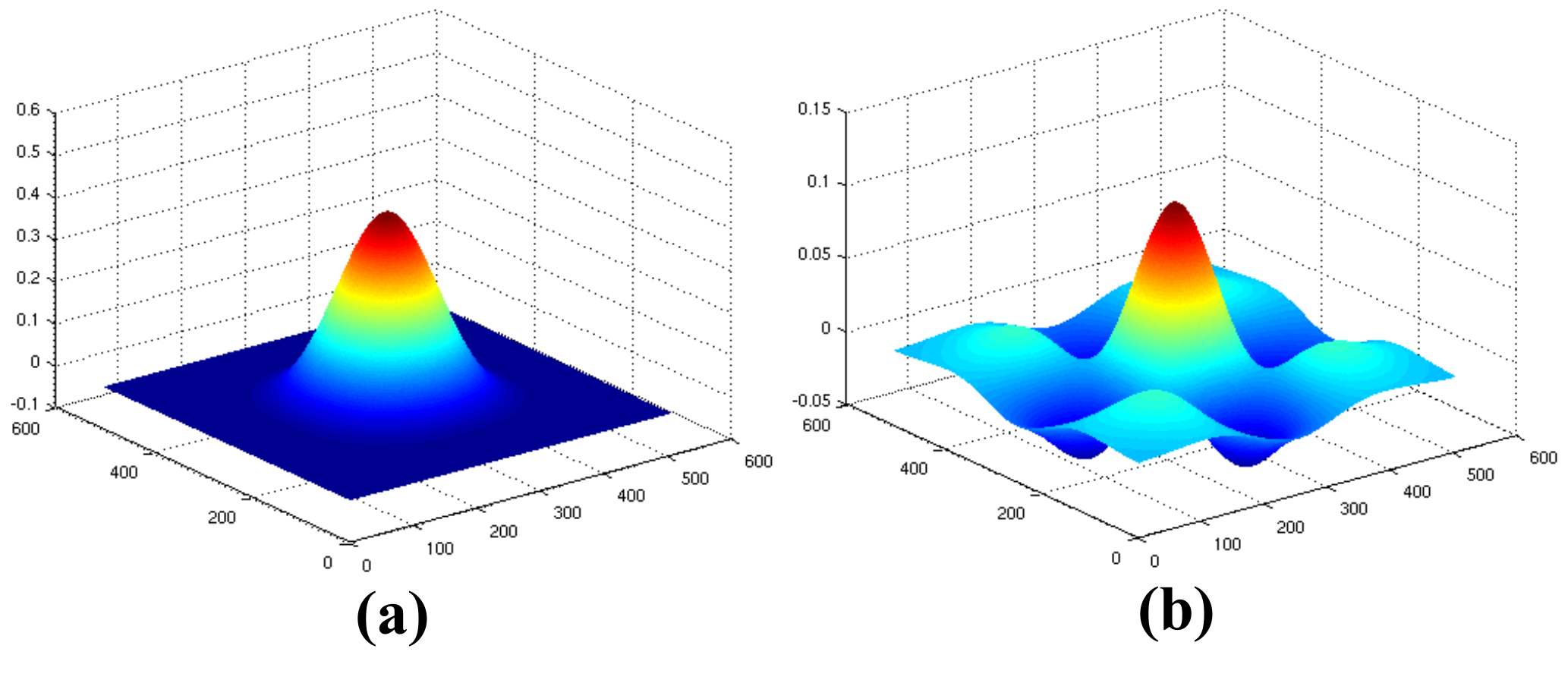}
\caption{2D starlet transform. (a) Two-dimensional scale function, $\phi$. 
(b) Two-dimensional wavelet, $\psi$.}
\label{FIG1}
\end{figure*}

Two-dimensional starlet transform is constructed from scale ($\phi$) and 
wavelet ($\psi$) functions\cite{STARCK2010} (Fig. \ref{FIG1}), given on Eqs. 
\ref{EQ1} and \ref{EQ2}:

\begin{center}
\begin{eqnarray}
\phi_{1D}(t) = \frac{1}{12}(|t-2|^3-4|t-1|^3+6|t|^3-4|t+1|^3+|t+2|^3) \nonumber \\
\phi(x,y) = \phi_{1D}(x)\phi_{1D}(y) \label{EQ1} \\
\frac{1}{4}\psi(\frac{x}{2},\frac{y}{2}) = \phi(x,y)-\frac{1}{4}\phi(\frac{x}{2},\frac{y}{2}) \label{EQ2}
\end{eqnarray}
\end{center}

\noindent where $\phi_{1D}$ is the one-dimensional B-spline of order 3 (B3-spline). Starck 
and Murtagh\cite{STARCK2006,STARCK2010,STARCK2011} used extensively the 
B3-spline as $\phi$, by its attributes: it is a smooth function, adequate to 
the isolation of larger structures in an image, and supports separability, 
allowing fast computation.

The pair of filters $(h,g)$ related to this wavelet is\cite{STARCK2010}:

\begin{eqnarray}
h_{1D}[k] = [\begin{array}{ccccc} 1 & 4 & 6 & 4 & 1\end{array}]/16, k = -2,...2 \\ \label{H1D}
h[k,l] = h_{1D}[k]h_{1D}[l] \label{H2D} \\
g[k,l] = \delta[k,l]-h[k,l] \label{EQ3}
\end{eqnarray}

\noindent where $\delta$ is defined as $\delta[0,0] = 1$, $\delta[k,l] = 0$, 
for $[k,l] \ne [0,0]$. From the structure of Eq. (\ref{EQ3}), one can see that 
starlet wavelet coefficients are achieved by the difference between two 
resolutions.

Starlet application is given by a convolution of an input image $c_{0}$ with 
the scale function $\phi$ (Eq. (\ref{EQ1})). Use of two-dimensional B3-spline 
as $\phi$ (Eq. \ref{H2D}) is given by a discrete convolution between the input 
image and the finite impulse response filter\cite{STARCK2010}:

\begin{center}
\begin{equation}
h = \left[\begin{array}{ccccc} \frac{1}{256} & \frac{1}{64} & \frac{3}{128} & \frac{1}{64} & \frac{1}{256} \\
\frac{1}{64} & \frac{1}{16} & \frac{3}{32} & \frac{1}{16} & \frac{1}{64} \\ 
\frac{3}{128} & \frac{3}{32} & \frac{9}{64} & \frac{3}{32} & \frac{3}{128} \\
\frac{1}{64} & \frac{1}{16} & \frac{3}{32} & \frac{1}{16} & \frac{1}{64} \\ 
\frac{1}{256} & \frac{1}{64} & \frac{3}{128} & \frac{1}{64} & \frac{1}{256} \end{array}\right]
\end{equation}
\end{center}

After $h$ and $c_{j}[k,l]$ convolution, wavelet coefficients 
are obtained from the difference $w_{j} = c_{j-1}-c_{j}$ (Eq. (\ref{EQ3})).

\subsection{Overview of the algorithms}

The proposed segmentation method is defined as follows:
\begin{itemize}
\item Starlet is applied in an input image $c_{0}$ , resulting in $L$ detail
levels: $D_1 , D_2 , \ldots, D_L$ , where $L$ is the last desired resolution
level.
\item First and second detail levels ($D_1$ and $D_2$ , respectively)
are assumed as noise and discarded.
\item TThird to L detail levels are summed ($D_3 + \ldots + D_L$).
\item Input image $c_{0}$ is subtracted from the result (Eq. (\ref{EQ4})).
Equation \ref{EQ4} describes these operations:
\end{itemize}

\begin{equation}
S = \sum_{i=3}^{L} D_{i}-c_{0} \label{EQ4},
\end{equation}

\noindent where $S$ is the image which represents the obtained
nanoparticles, $\sum{D_i}$ represents the sum of $D_3$ to $D_L$,
and $c_0$ is the input image.

For a clear overview of the proposed method, its pseudocode is listed in 
Algorithm \ref{ALG1}.

\begin{algorithm} 
\SetAlgoNoEnd
\DontPrintSemicolon 
\KwIn{A grayscale image, $c_{0}$.\\
Number of resolutions to be calculated, $L$.}
\KwOut{Detail coefficients from starlet transform, $w_{j}$.\\
A image that presents the nanoparticles contained in the
original image, $imgnp$.}
\textbf{mirroring(}$c_{0}$\textbf{)};\;
\For{$j \gets 1$ \textbf{to} $L$} {
	$h \gets$ \textbf{hgen(}j\textbf{)};\;
	$c_{j} \gets$ \textbf{convolution(}$c_{j-1},h$\textbf{)};\;
	$w_{j} \gets c_{j-1}-c_{j}$;\;
	\textbf{unmirroring(}$c_{j}$\textbf{)};\;
	\textbf{increment(}$j$\textbf{)};\;
}
\textbf{initialize} $sum$ \textbf{to} 0;\;
\For{$j \gets 3$ \textbf{to} $L$} {
	$sum \gets sum + w_{j};$
}
$imgnp \gets sum - c_{0};$

\Return{$w_{j}$, $imgnp$}\;
\caption{Pseudocode for determination of nanoparticles in an image, based 
on starlet algorithm application (adapted from {[Starck et al., 2011]}).}
\label{ALG1}
\end{algorithm}

$W = \{w_{1}, \cdots, w_{L}, c_{L}\}$ represents the input image starlet transform. 
Function \textbf{hgen()} (Algorithm \ref{ALG2}), referenced on Algorithm 
\ref{ALG1}, is applied when $j$ is incremented; so for $j>1$, $h$ has $2^{j-1}$ 
zeros between its elements. Algorithm \ref{ALG2} is also responsible for the generation of $h$.

\begin{algorithm} 
\SetAlgoNoEnd
\DontPrintSemicolon 
\KwIn{$h_{1D}$ filter, given by Eq. (\ref{H1D}).\\
Current resolution level, $j$.}
\KwOut{Filter $h_{2D}$, $h$.}
\If{$j = 0$}{
	$h \gets h_{1D}$;\;
}
\Else{
	$M \gets$ \textbf{size(}$h_{1D},2$\textbf{)};\;
	\textbf{initialize(}$k$ \textbf{to} $0$\textbf{)};\;
	\For{$i \gets 1$ \textbf{step} $2^{i-1}$ \textbf{to} $M+2^{i-1}*(M-1)$}{
		\textbf{increment(}$k$\textbf{)};\;
		$h(i) \gets h_{1D}(k)$;\;
	}
}
\textbf{initialize} $aux$ \textbf{to} 0;\;
$aux \gets sum(sum(h'*h))$;\;
$h \gets (h'*h)/aux$;\;
\Return{$h$}\;
\caption{\textbf{hgen:} $h$ generation and zero-inserting after each interaction.}
\label{ALG2}
\end{algorithm}

\section{Evaluation} 
\label{EVALUATION}

\subsection{Image dataset}

A data set consisting of $30$ images was employed in order to evaluate the 
proposed algorithms. These images were obtained from natural rubber samples 
with gold nanoparticles using scanning electron microscopy.

Gold nanoparticles were reduced in natural rubber at different time 
periods: $6$, $9$, $15$ and $30$ minutes. SEM images were obtained 
in magnifications of $100,000$ and $200,000$ times. 
More details about NR/Au samples are given by Barboza-Filho et al \cite{BARBOZAFILHO2012}.

SEM measurements were carried out using a FEI Quanta 200 FEG microscope with field
emission gun (filament), equipped with a large field detector (LFD), Everhart-Thornley secondary
electron detector, and a solid state backscattering detector and pressure of 1.00 Torr aprox. (low
vacuum), as well as uncoated surface. The images in the data set were obtained by secondary detector,
due to more resolution/specificity.

\subsection{Evaluation method}

Precision, recall and accuracy measures\cite{OLSON2008,WANG2013} 
were employed to evaluate the performance of the proposed method. These 
measures are based on the concepts of true positives (TP), true negatives 
(TN), false positives (FP) and false negatives (FN).

In comparison to the ground truth of an input image,
\begin{itemize}
\item TP are pixels correctly labeled as gold nanoparticles;
\item FP are pixels incorrectly labeled as gold nanoparticles;
\item FN are pixels incorrectly labeled as background;
\item TN are pixels correctly labeled as background.
\end{itemize}

Based on these assertions, precision, recall and accuracy are defined as:

\begin{eqnarray}
precision &=& \frac{TP}{TP+FP} \nonumber \\
recall &=& \frac{TP}{TP+FN} \nonumber \\
accuracy &=& \frac{TP+TN}{TP+TN+FP+FN} \nonumber
\end{eqnarray}

Precision expresses retrieved pixels that are relevant, while recall expresses relevant pixels that
were retrieved. Accuracy, on the other hand, means the proportion of true retrieved results.

\section{Experimental results} 
\label{EXPERIMENTALRESULTS}

\begin{figure*}[htb]
\centering
\includegraphics[width=15cm]{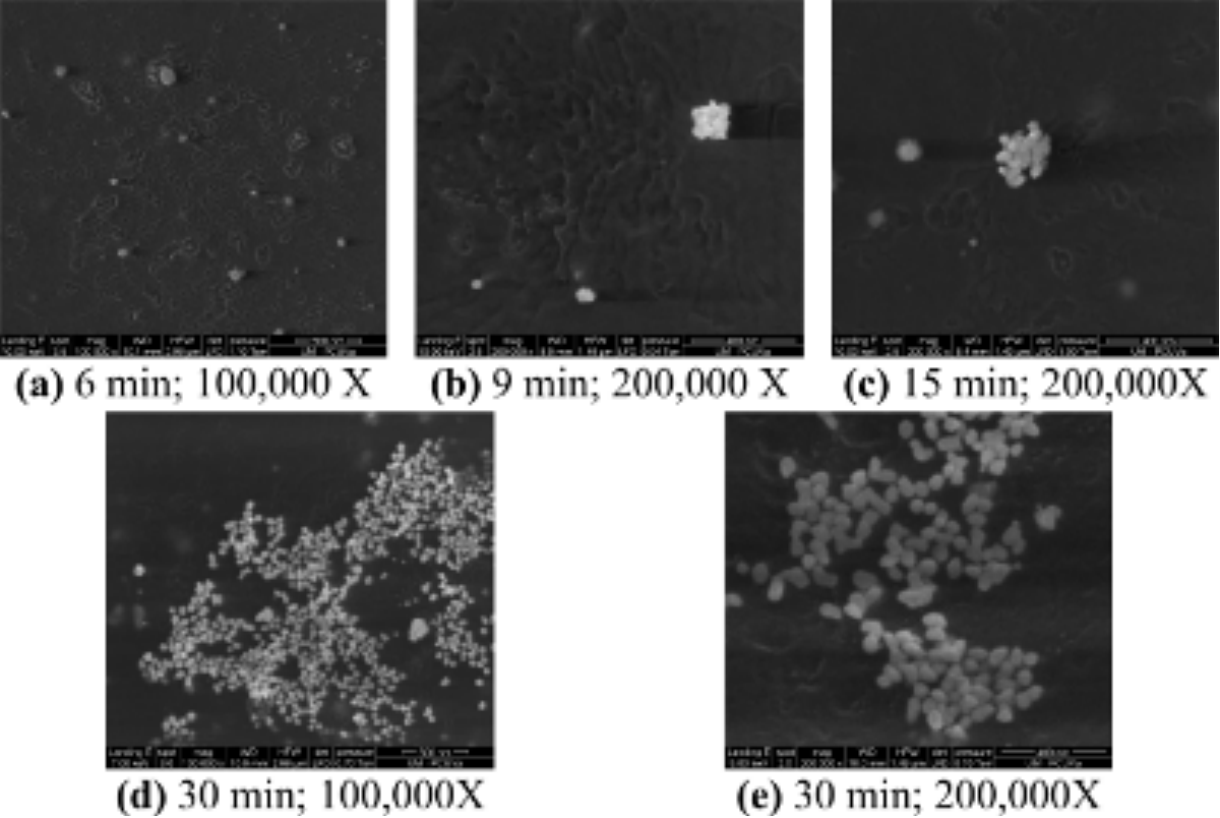}
\caption{Scanning electron microscopy images obtained from NR/Au samples. 
Reduction time of gold nanoparticles: from $6$ to $30$ minutes. 
Magnification: $100,000$ X and $200,000$ X.}
\label{FIG2}
\end{figure*}

To introduce the results obtained with the proposed method, five images, 
which belong to the data set, are presented with different distribution, 
nanoparticle amount and size (Fig. \ref{FIG2}). The lighter regions of 
the images correspond to gold nanoparticles in natural rubber surface.

The proposed method was applied in the test images with $L = 3$ to $L = 10$, 
and precision, recall and accuracy were obtained for each result.

\begin{figure*}[hbt] 
\centering
	\subfloat[Precision.]{\label{FIG3a}{\includegraphics[width=0.33\textwidth]{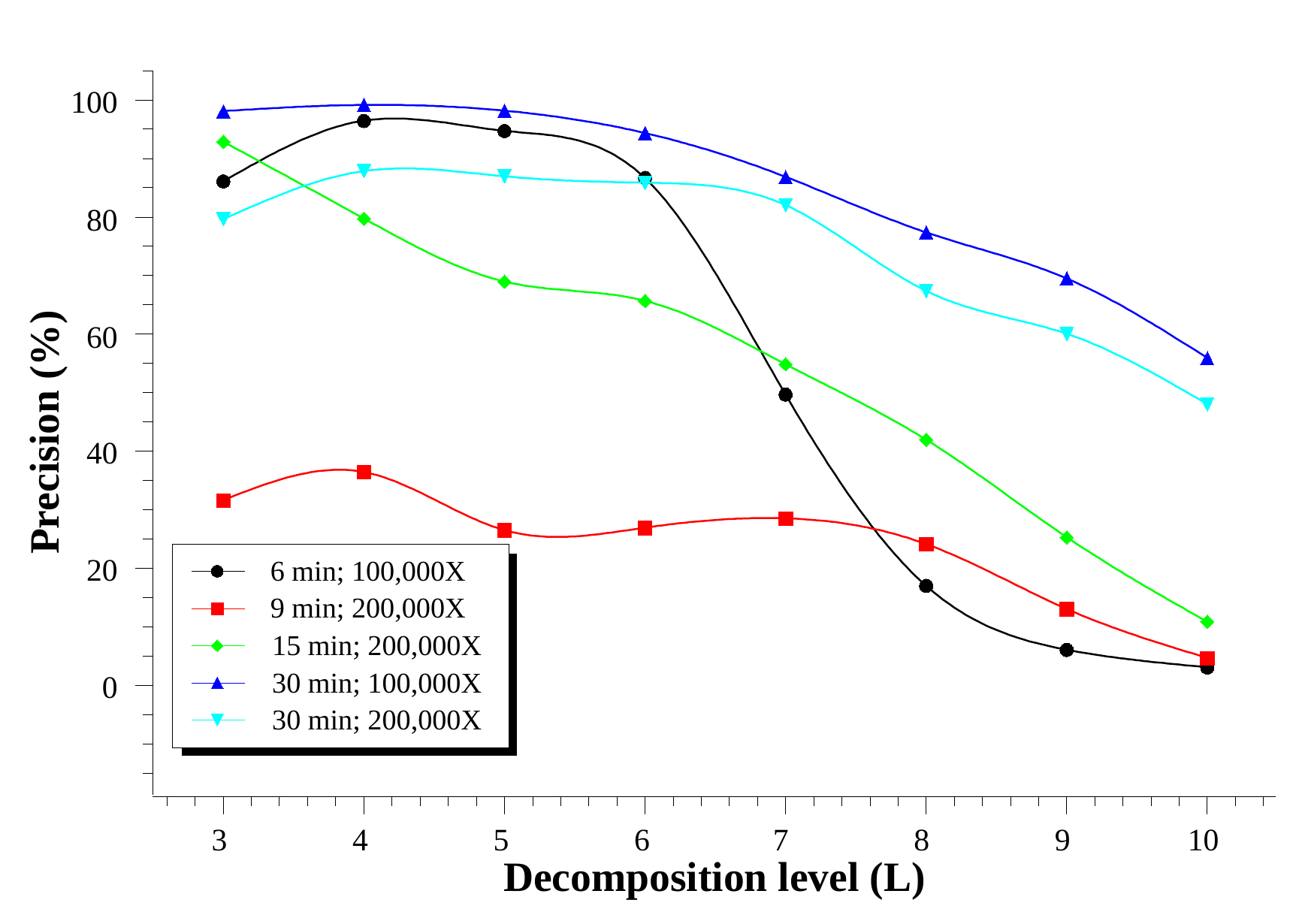}}}\hfill
	\subfloat[Recall.]{\label{FIG3b}{\includegraphics[width=0.33\textwidth]{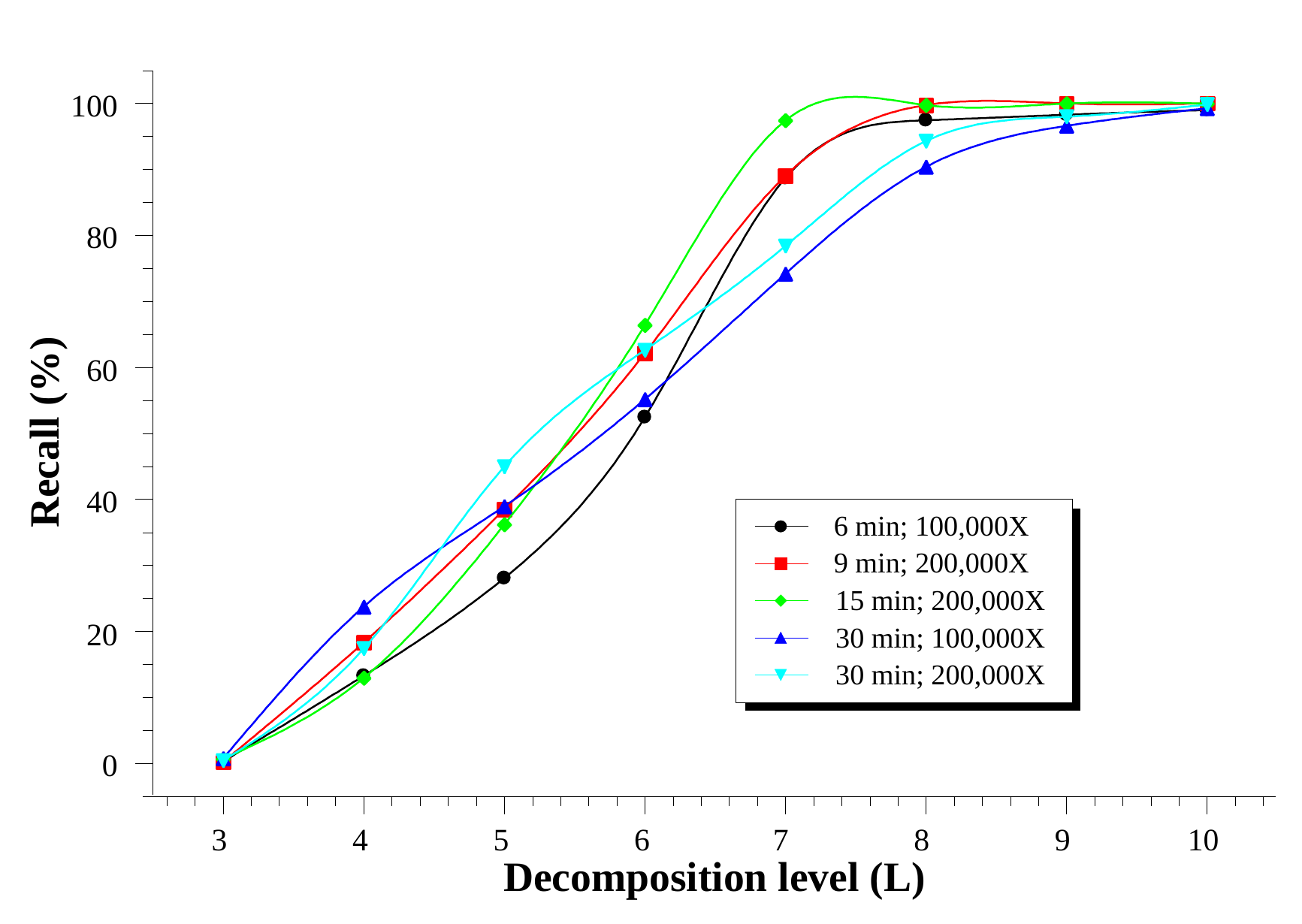}}}\hfill
	\subfloat[Accuracy.]{\label{FIG3c}{\includegraphics[width=0.33\textwidth]{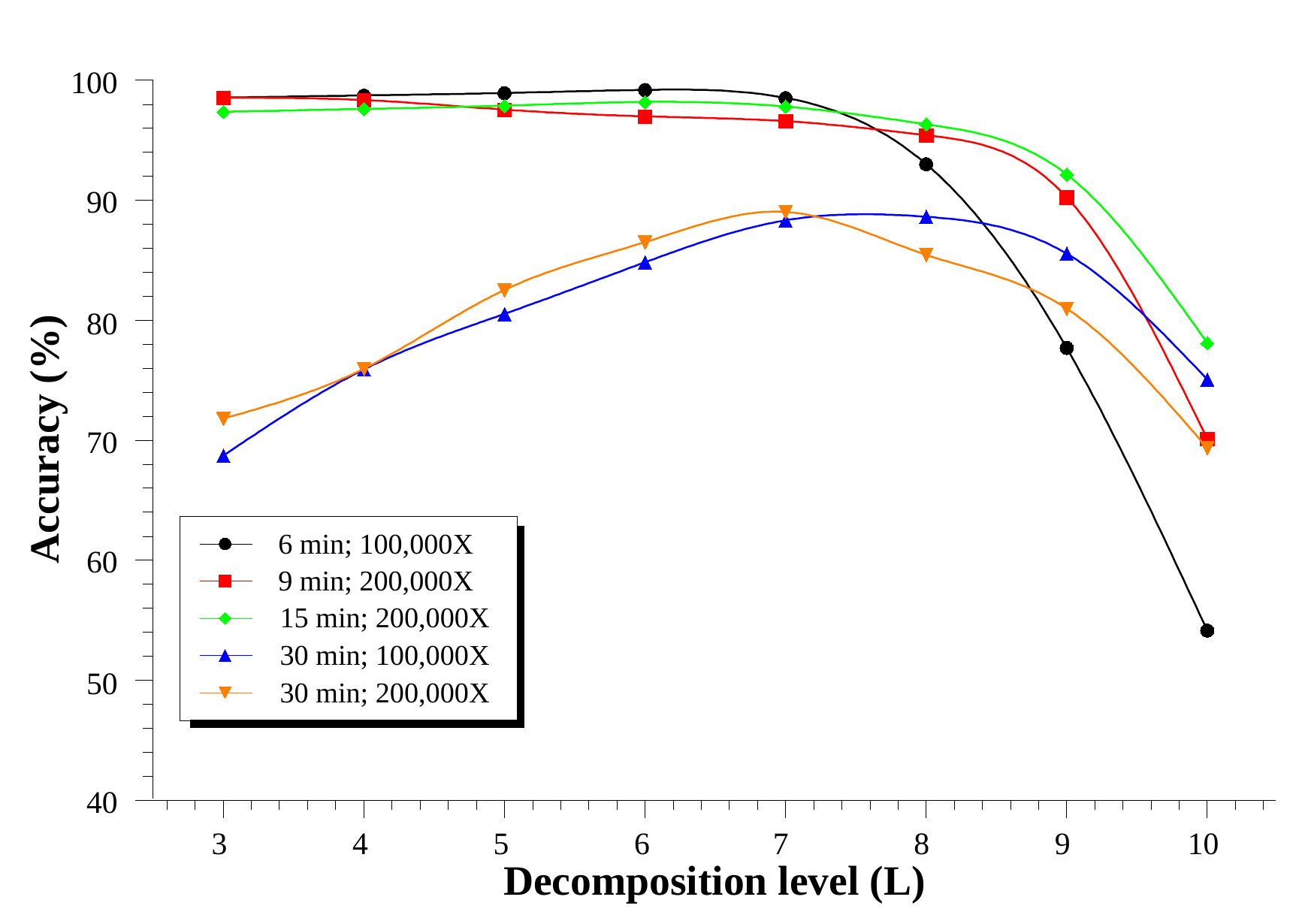}}}
\caption{Precision, recall and accuracy values for Fig. \ref{FIG2} images.}
\label{FIG3}
\end{figure*}

One could see from Fig. 3 that the method accuracy, in general, increases 
until $L = 6$ and starts to decrease when $L = 7$. Precision varies widely 
(between $30\%$ and $100\%$ for $L = 3$), but decreases for all images 
after $L = 7$. However, recall has a similar behavior for these images, 
as $L$ increases.

For a satisfactory segmentation degree, an optimal relationship between 
$FN$ and $FP$ pixels (i.e. precision and recall) becomes necessary. Optimal $L$ 
levels were determined based on high accuracy, precision and recall, in 
order to automatically choose the best result for each image:

\begin{itemize}
\item\textbf{6 min; 100,000X:} from Fig. \ref{FIG3}, greater accuracy 
levels are given for $L$ equals to $5$, $6$ and $7$. $L = 6$ (precision = $86,64\%$; 
recall = $52,42\%$; accuracy = $99,19\%$) was chosen because $L = 5$ has a 
low recall ($28,06\%$) and $L = 7$ has a low precision ($49,67\%$), although 
$L = 6$ and $L = 7$ have similar accuracy ($98,52\%$ for $L = 7$).

\item\textbf{9 min; 200,000X:} although Fig. \ref{FIG3} presents a 
greater accuracy level for $L = 3$, the use of this level is not appropriate, 
since its recall is almost null. Other levels to be considered are $L$ 
equals to $6$ and $7$. $L = 7$ (precision = $28,49\%$; recall = $89,02\%$; 
accuracy = $96,58\%$) was chosen because $L = 6$ has lower precision ($26,89\%$) 
and recall ($62,13\%$).

\item\textbf{15 min; 200,000X:} greater accuracy levels are given for $L$ 
equals to $5$, $6$ and $7$, according to Fig. \ref{FIG3}. Even with $L = 6$ 
presenting higher accuracy ($98,18\%$), $L = 7$ (precision = $54,82\%$; 
recall = $97,39\%$; accuracy = $97,80\%$) was chosen by a higher recall 
value (for $L = 6$, $66,38\%$).

\item\textbf{30 min; 100,000X:} also from Fig. \ref{FIG3}, greater accuracy 
values are given for $L$ equals to $7$, $8$ and $9$. $L = 8$ (precision = $77,36\%$; 
recall = $90,39\%$; accuracy = $88,31\%$) has a greater accuracy value, 
precision higher than $L = 9$ ($69,47\%$) and recall higher than $L = 7$ ($74,18\%$).

\item\textbf{30 min; 200,000X:} finally, greater accuracy levels are given 
for $L$ equals to $6$, $7$ and $8$, from Fig. \ref{FIG3}. $L = 7$ (precision = $81,99\%$, 
recall = $78,39\%$, accuracy = $86,49\%$) was chosen for having greater 
accuracy, precision higher than $L = 8$ ($67,34\%$) and recall higher than 
$L = 6$ ($62,61\%$).
\end{itemize}

\begin{figure*}[hbt]
\centering
\includegraphics[width=10cm]{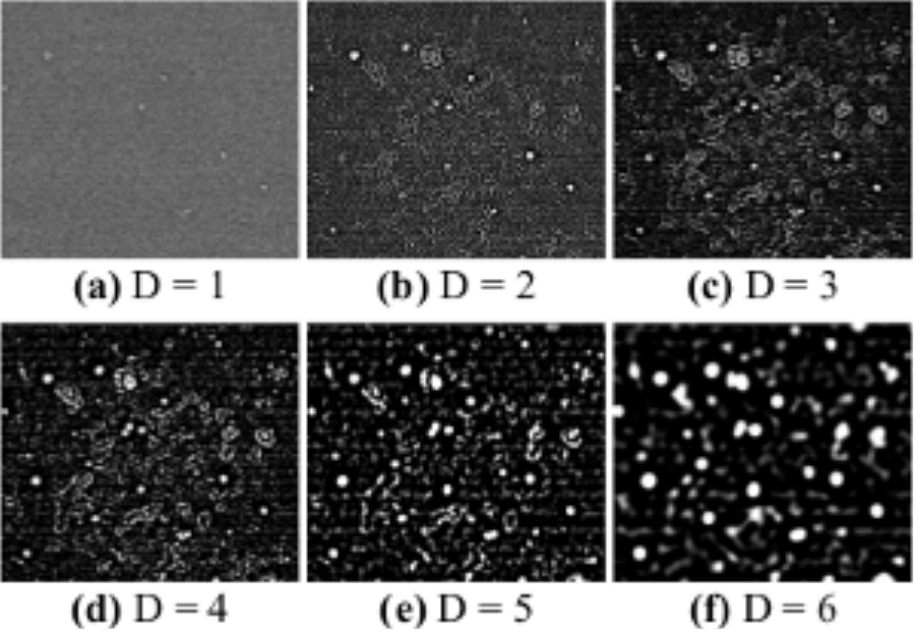}
\caption{Starlet detail decomposition levels of Fig. \ref{FIG2} (a).
$D=1$ and $D=2$ (Fig. \ref{FIG4} (a), (b)) were discarded from algorithm 
application due to large amount of noise. Higher levels shows nanoparticle 
locations more clearly; however, background regions tend to aggregate, 
reducing algorithm accuracy.}
\label{FIG4}
\end{figure*}

$L = 6$, chosen for Fig. \ref{FIG2} (a), results in six decomposition detail 
levels. Starlet detail decomposition levels of Fig. \ref{FIG2} (a), from 
1 to 6, are presented in Fig. \ref{FIG4}.

Higher detail decomposition levels emphasize the sample surface. The 
first detail level (Fig. \ref{FIG4} (a)), D1, presents mostly noise; the 
second level (Fig. \ref{FIG4} (b)), D2, also presents large noise amount. 
Smoothing factor grows as the decomposition level increases. Higher detail 
levels ((Fig. \ref{FIG4} (c) to \ref{FIG4} (f)) represents the sample 
surface with greater precision when the noise decreases, although regions tend
to aggregate according to the increasing of detail level.

\begin{figure*}[htb]
\centering
\includegraphics[width=15cm]{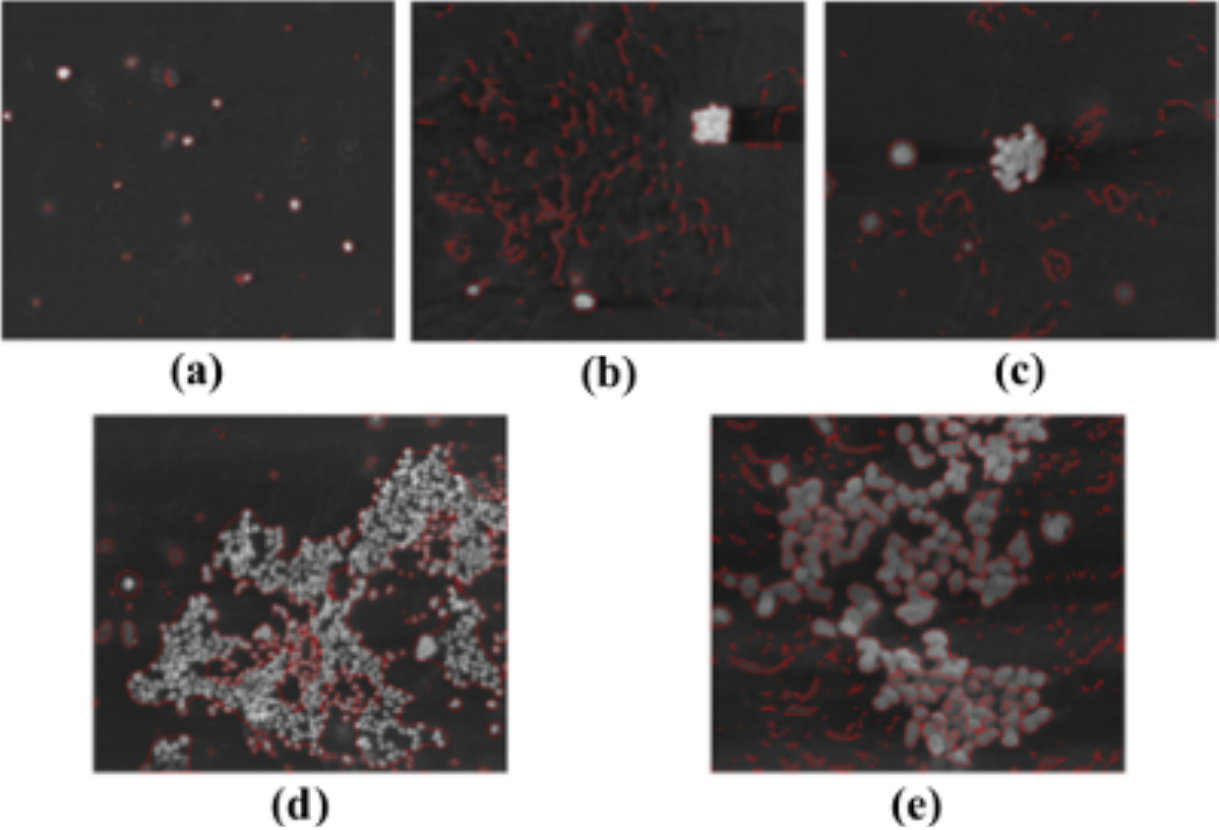}
\caption{Results of Algorithm 1 applied in Fig. \ref{FIG2}. Nanoparticles 
recognized by the proposed algorithm appear highlighted.}
\label{FIG5}
\end{figure*}

After starlet application, decomposition levels from $3$ to $6$ (Fig. \ref{FIG4} (c) 
to \ref{FIG4} (f)) are added and subtracted from the original image (Fig. \ref{FIG2} (a)). 
The result of Algorithm 1 applied in Fig. \ref{FIG2} (a) is shown in Fig. \ref{FIG5} (a). 
Similarly, results of Algorithm 1 applied in Fig. \ref{FIG2} (b) to (e) 
are shown in Fig. \ref{FIG5} (b) to (e).

Ground truth (GT) images obtained from Fig. \ref{FIG2} are used to evaluate 
the performance of the proposed algorithm. These images were acquired
manually by a specialist, using GIMP\footnote{Available freely at
\href{www.gimp.org}{www.gimp.org}.}, an open source graphics software.

\begin{figure}[htb]
\centering
\includegraphics[width=10cm]{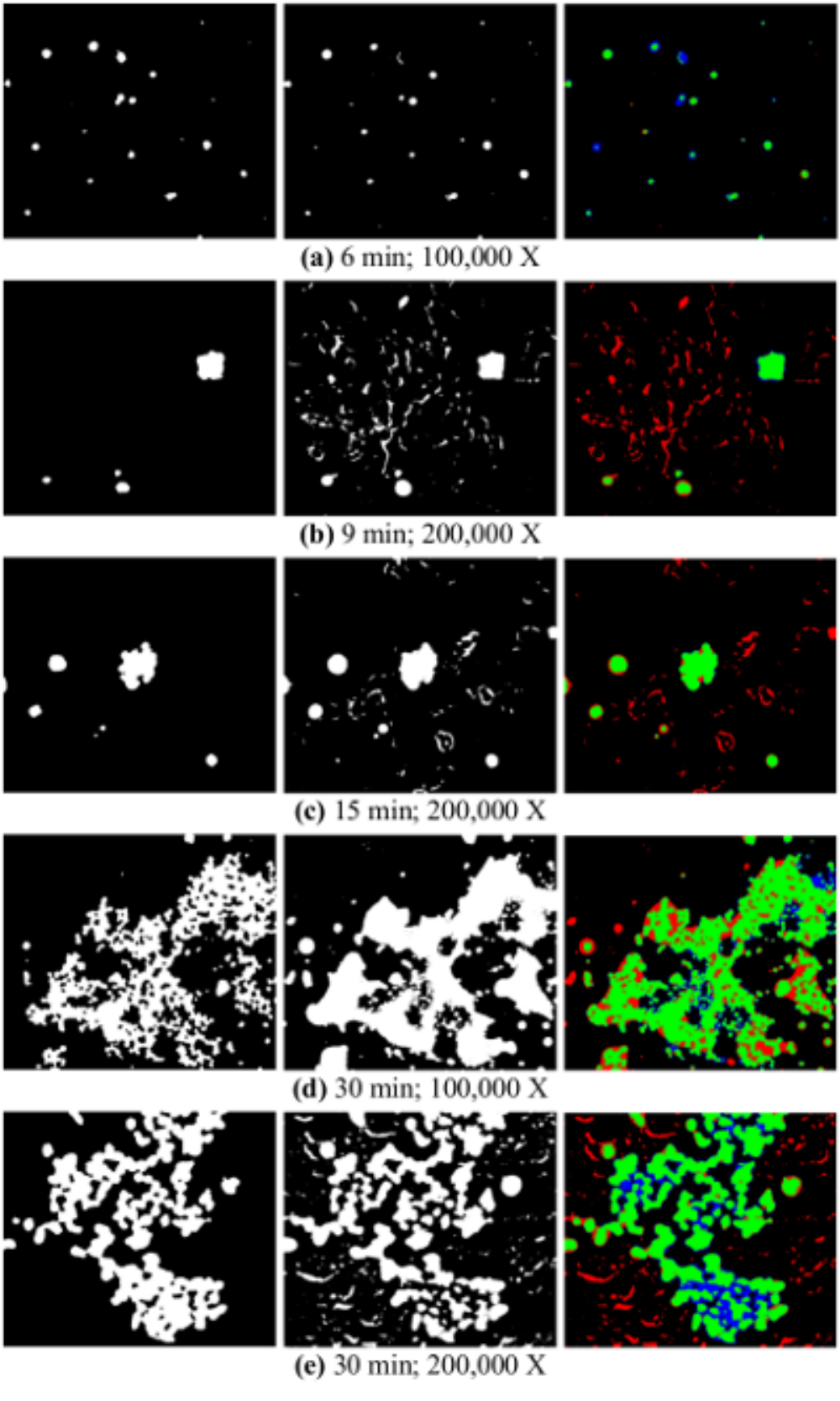}
\caption{Ground Truth (first column), result of Algorithm 1 application 
(second column) and comparison (third column) between GT and results. 
Green: true positive (TP) pixels; blue: false negative (FN) pixels; red: 
false positive (FP) pixels.}
\label{FIG6}
\end{figure}

Results of Algorithm 1 
applied in Fig. \ref{FIG2} are represented by binary images, where nanoparticles 
(the region of interest) are white regions and background by black 
regions. For comparison effects, green is assigned to true positive (TP) 
pixels, blue to false negative (FN) pixels and red to false positive (FP) 
pixels (Fig. \ref{FIG6}).

Most gold nanoparticles shown in GT images were located by the proposed 
method. In some cases, nanoparticles were not completely encountered; this 
phenomenon is represented by green spots surrounded by blue pixels. False 
positive results (red pixels) appear mostly in background. This is given 
by high roughness in the surface of natural rubber samples.

\subsection{Inaccurate results}

Since the proposed method is aimed at boundary-based segmentation, input 
images with high natural rubber roughness (as presented in Fig. \ref{FIG6} (b) and (e)) 
presents higher FP values, which leads to lower algorithm accuracy.
SEM images obtained by backscattering detectors do not show surface
topology, possibly minimizing this issue.
Likewise, a  large amount of aggregated nanoparticles leads to a high FN
value (Fig. \ref{FIG6} (d)).

\section{Conclusion} 
\label{CONCLUSION}

In this study we present a method for segment scanning electron microscopy images based on
starlet wavelets. This algorithm uses starlet decomposition detail levels to determine the edges of
objects within an input image. After obtaining the starlet detail levels, the higher detail levels are added
and background is removed from this result. Therefore, only the desired details are shown.

An application of the method is shown in images obtained from natural rubber samples with
gold nanoparticles. In this application, our method obtained accuracy higher than 85\% in images
obtained by secondary detectors.

There are some issues concerning the presented method. Structural details of the input image
can interfere in the final result, being labeled as desired areas. However, the method presented high
accuracy for all dataset images. For better results, we suggest application in SEM images obtained by
backscattering detectors, that do not show the surface topology.

Results given by this algorithm will be used in future studies, to computationally estimate the
density distribution of gold nanoparticles in natural rubber samples, and also to predict reduction
kinetics of gold nanoparticles at different time periods.

\section*{Acknowledgements} 
\label{ACKNOWLEDGEMENTS}

The authors would like to acknowledge the Brazilian foundations of research 
assistance CNPq, CAPES and FAPESP. This research is supported by FAPESP 
(Procs 2010/03282-9 and. 2011/09438-3).

\clearpage

\bibliographystyle{ieeetr}
\bibliography{references}

\end{document}